\documentclass[sigconf]{acmart}

\usepackage{booktabs} 
\usepackage{amsmath}
\usepackage{graphicx}
\usepackage{colortbl}
\usepackage{mathtools}
\usepackage{url}
\usepackage{multirow}

\DeclareMathOperator*{\argmin}{\arg\!\min}

\setcopyright{rightsretained}

\begin{document}
\title{Protecting Sensory Data against Sensitive Inferences}

\author{Mohammad Malekzadeh}
\affiliation{ \institution{Queen Mary University of London}
\city{London}
\country{UK}}
\email{m.malekzadeh@qmul.ac.uk}

\author{Richard G. Clegg}
\affiliation{ \institution{Queen Mary University of London} 
\city{London}
\country{UK}}
\email{r.clegg@qmul.ac.uk}

\author{Andrea Cavallaro}
\affiliation{ \institution{Queen Mary University of London}
\city{London}
\country{UK}}
\email{a.cavallaro@qmul.ac.uk}

\author{Hamed Haddadi}
\affiliation{ \institution{Imperial College London}
\city{London}
\country{UK}}
\email{h.haddadi@imperial.ac.uk}



\copyrightyear{2018} 
\acmYear{2018} 
\setcopyright{acmlicensed}
\acmConference[W-P2DS'18]{1st Workshop on Privacy by Design in Distributed Systems }{April 23--26, 2018}{Porto, Portugal}
\acmBooktitle{W-P2DS'18: 1st Workshop on Privacy by Design in Distributed Systems , April 23--26, 2018, Porto, Portugal}
\acmPrice{15.00}
\acmDOI{10.1145/3195258.3195260}
\acmISBN{978-1-4503-5654-1/18/04}

\begin{abstract}
	There is growing concern about how personal data are used when users grant applications direct access to the sensors of their mobile devices.  In fact, high resolution temporal data generated by motion sensors reflect directly the activities of a user and indirectly physical and demographic attributes. In this paper, we propose a feature learning architecture for mobile devices that provides flexible and negotiable privacy-preserving sensor data transmission by appropriately transforming raw sensor data. The objective is to move from the current binary setting of granting or not permission to an application, toward a model that allows users to grant each application permission over a limited range of inferences according to the provided services. The internal structure of each component of the proposed architecture can be flexibly changed and the trade-off between privacy and utility can be negotiated between the constraints of the user and the underlying application. We validated the proposed architecture in an activity recognition application using two real-world datasets, with the objective of recognizing an activity without disclosing gender as an example of private information. Results show that the proposed framework maintains the usefulness of the transformed data for activity recognition, with an average loss of only around three percentage points, while reducing the possibility of gender classification to around 50\%, the target random guess, from more than 90\% when using raw sensor data.  We also present and distribute  MotionSense, a new dataset for activity and attribute recognition collected from motion sensors.
\end{abstract}

\maketitle

\section{Introduction}

Smartphones and wearable devices are equipped with sensors such as accelerometers, gyroscope, barometer and light sensors that are directly accessed by applications (apps) to provide through a cloud service analysis and statistics about, for example, the activities of the user. However, by granting to these apps access to raw sensor data, users may unintentionally reveal information about gender, mood, personality, which is unnecessary for the specific services. 

\begin{figure}[t!]
	\centering
	\includegraphics[scale=0.34]{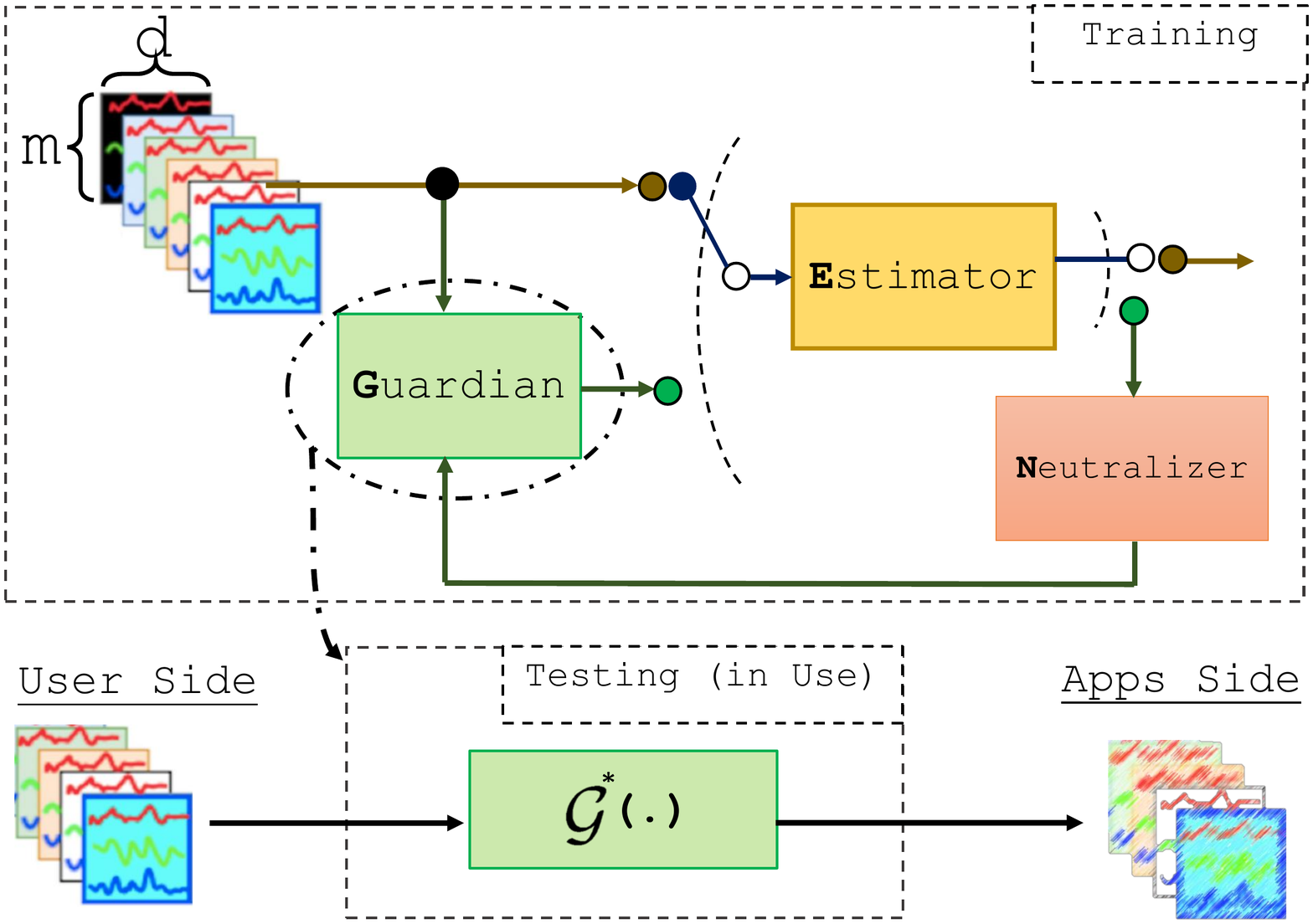}
	\caption{\label{fig:gen_arc} {GEN Architecture: First, the Estimator is trained; then the Guardian is trained using the Estimator with the help of the Neutralizer.}}
\end{figure}

To address this problem, we introduce the Guardian-Estimator-Neutralizer (GEN) framework that, instead of granting apps direct access to sensors, is designed to share only a transformed version of the sensor data, based on the functions and requirements of each application and privacy considerations. The \textit{Guardian} provides an inference-specific transformation, the \textit{Estimator} guides the Guardian by estimating sensitive and non-sensitive information in the transformed data, and the \textit{Neutralizer} is an optimizer that helps the Guardian converge to a near-optimal transformation function (see Figure \ref{fig:gen_arc}). 

Unlike privacy-preserving works that only hide users' identity by sharing population data using generative models for data synthesis~\cite{beaulieu2017privacy,huang2017context}, our solution concerns sensitive information included in a single user's data. There are, however, some methods which transform only selected temporal sections of sensor data that correspond to predefined sensitive activities~\cite{malekzadeh2017replacement, saleheen2016msieve}, our  framework enables concurrently eliminating private information from each section of data, while keeping the utility of shared data.

GEN is a feature learning and data reconstruction framework that helps to efficiently establish a trade-off between apps utility and user privacy. Specifically, in this paper, we instantiate the framework for an activity recognition application based on data recorded by the accelerometer and gyroscope of a smartphone. In the context of this application, we categorize information that can be inferred from sensor data into two types: information about a predefined set of activities of the user (\textit{non-sensitive inferences}) and information about attributes of the user such as gender, age, weight and height (\textit{sensitive inferences}).

Our goal is to establish a tradeoff between the ability of the apps to accurately infer non-sensitive information to maximize their \textit{utility} and the reduction of revealed sensitive information to minimize the risk of \textit{privacy} infringement. 
We show that GEN can accurately maintain the usefulness of the released (transformed) data for activity recognition while considerably reducing the risk of attribute recognition.\footnote{The code and data used in this paper are publicly available at:\\ https://github.com/mmalekzadeh/motion-sense}\\

\section{Problem Definition}

Let $X(t) = \big(X_1(t), X_2(t), \dots , X_m(t)\big)$ be the recorded values of the $m$ sensor-data components during a collection period of duration $T$, where $t \in \{1, 2, \dots ,T\}$. We assume the data to be synchronized and collected at the same frequency.

Let us consider a running window of duration $d$ that contains consecutive values of $X(t)$ from time $t$ to $t+d-1$. Let $S_d(t)$ be the corresponding section of the time-series:
$$
	S_d(t)= X[t,t+d-1] = \Big(X(t), X(t+1), \dots ,X(t+d-1) \Big),
$$
where the value of $d$ should be chosen such that the running window be large enough for making desired inferences by apps. However, in order to be computationally effective, it should not be chosen very large. For simplicity, we remove the index $t$, from $S_d(t)$, in the following.

We define two types of inference on each $S_d$:  inference of \textbf{s}ensitive information, $I_{\textbf{s}}(.)$, and inference of \textbf{n}on-sensitive information, $I_{\textbf{n}}(.)$. 
Our goal is to find a transformation function, $\mathcal{G}^{*}(.)$, in a way that the transformed data
$\hat{S}_d^{*}=\mathcal{G}^{*}(S_d)$ are such that $I_{\textbf{s}}(\hat{S}_d^{*})$ fails to reveal private information, whereas $I_{\textbf{n}}(\hat{S}_d^{*})$ generates inferences that are as accurate as $I_{\textbf{n}}(S_d)$. Here, $\hat{S}_d$ is the transformation of corresponding ${S}_d$, and  $\hat{S}_d^{*}$ is its optimal privacy-preserving transformation.\\
\section{Learning the Inference-Specific Transformation}
\begin{figure}[t!]
	\centering
	\includegraphics[scale=0.3]{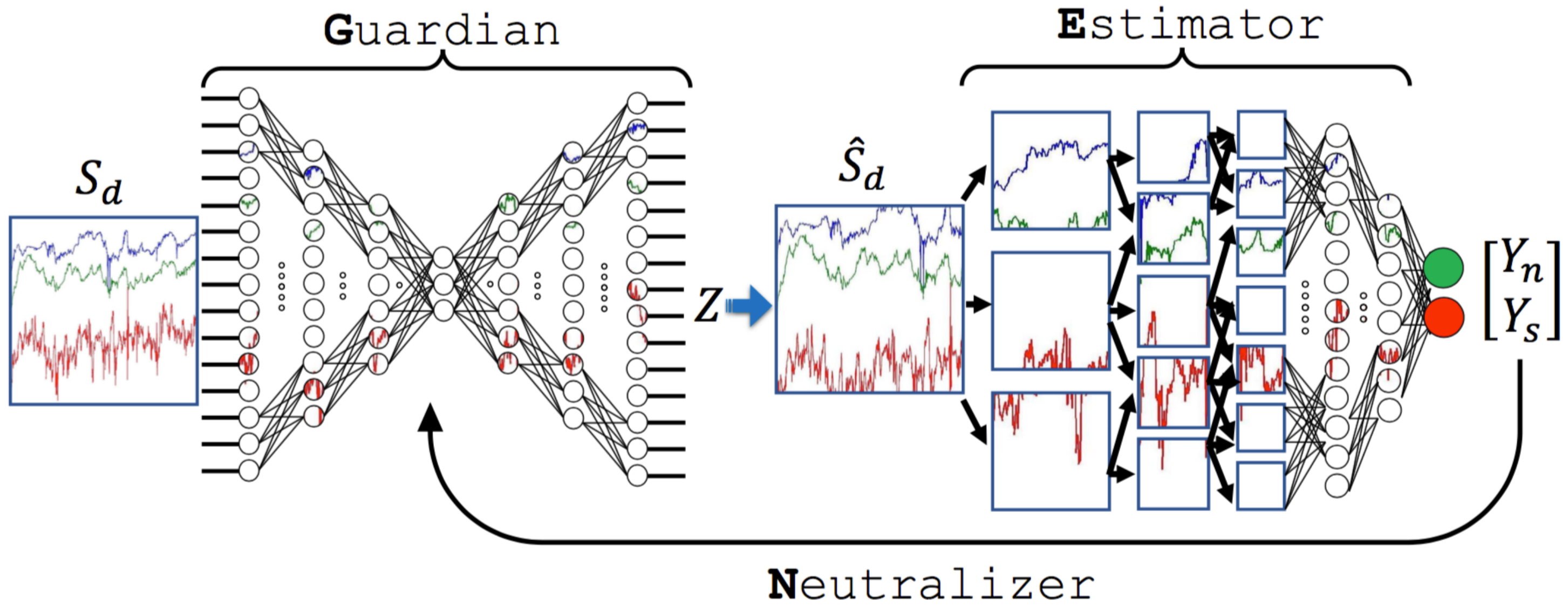}
	\caption{{\label{fig:aen_cnn} An instantiation of GEN for activity recognition from sensor data without revealing the gender information. The Guardian is an \textit{autoencoder}. The Estimator is a \textit{multi-task ConvNet}.}}
\end{figure}

We present the proposed framework that includes three  components: the Guardian, the Estimator, and the Neutralizer~(Figure~\ref{fig:gen_arc}), and discuss its instantiation for an activity recognition application~(Figure~\ref{fig:aen_cnn}).
\\

The \textbf{Guardian}, which provides \textit{inference-specific transformation}, is a feature learning framework that recognizes and distinguishes discerning features from data. In the specific implementation of this paper, we use a \textit{deep autoencoder}~\cite{vincent2008extracting}  as Guardian. An autoencoder is a neural network that tries to reconstruct its input based on an objective function. Here, the autoencoder receives a section of \textit{m-dimensional} time-series with length of $d$ as input, and produces a time-series with the same dimensionality as the output; based on the Neutralizer's objective function, which is described below.
\\
	
The \textbf{Estimator} quantifies how accurate an algorithm can be at making sensitive and non-sensitive inferences on the transformed data. In the specific implementation of this paper, we use a \textit{multi-task convolutional neural network (MTCNN)} as Estimator~\cite{yang2015deep}. The shape of input is similar to the Guardian and the shape of output depends on the number of activity classes.
 MTCNN has the ability to share learned representations from input between several tasks. More precisely, we try to simultaneously optimize a CNN with two types of loss function, one for sensitive inferences and another for non-sensitive ones. Consequently, MTCNN will learn more generic features, which should be used for several tasks, at its earlier layers. Then, subsequent layers, which become progressively more specific to the details of the desired task, can be divided into multiple branches, each for a specific task.
 \\

The \textbf{Neutralizer}, the most important contribution of this paper, is an optimizer that helps the Guardian find the optimal $\mathcal{G}^{*}(\cdot)$ for transforming each section $S_d$ into $\hat{S}^{*}_d$ using as objective
$$
{\mathcal{G}}^{*}(.) = \argmin_{\mathcal{G}(.)\in\mathcal{F}}  \Bigg( p\Big(I_{\textbf{s}}\big(\mathcal{G}({S}_d)\big)\Big) - p\Big(I_{\textbf{n}}\big(\mathcal{G}({S}_d)\big)\Big) \Bigg),
$$
where $p\big(I_s(\cdot)\big)$ and $p\big(I_n(\cdot)\big)$ are the probabilities of making sensitive and non-sensitive inferences, respectively, and the $\mathcal{F}$ is the set of all possible transformation functions for the Guardian. In the specific application of this paper the Neutralizer is a multi-task objective function used by backpropagation to update the weights of the Guardian (autoencoder). The $\mathcal{F}$ is also the set of all possible weight matrices for the selected autoencoder.

Particularly, we aim to transform each section $S_d$ such that we can recognize an activity from  $\hat{S_d}$ without revealing the gender of the user. For each section $S_d$, let $Y_{\textbf{a}}(S_d)$ and $Y_{\textbf{a}}(\hat{S_d})$ be the true and predicted class of activity, respectively, and $Y_{\textbf{g}}(\hat{S}_d)$ be the predicted gender class. We define the Neutralizer's objective function as
\\
\begin{multline}\label{eq:neutralizer}
  		\hat{S}^{*}_d = 
  		\argmin_{\hat{S}_d} \Big(\mid{(0.5 - Y_{\textbf{g}}(\hat{S}_d)})\mid  - \sum_{i=1}^{c} - Y_{\textbf{a}}^{i}({S}_d)\log{Y_{\textbf{a}}^{i}(\hat{S}_d)} \Big),
\end{multline}
\\
where $c$ is the number of activity classes. In the r.h.s.~of the equation, the first part is our custom gender-neutralizer loss function  and the second part is a categorical cross entropy. The constant $0.5$ is the desired confidence for a gender predictor that will process the transformed data.\\

\section{Experiments} \label{sec:exp}
We validate the proposed framework on recognizing the following activities from smartphone motion sensors: \textit{Downstairs, Upstairs, Walking, Jogging}. The  non-sensitive inferences, $ I_{\textbf{n}}$, is the recognition of the activities, whereas the sensitive inference, $I_{\textbf{s}}$, is the recognition of gender.

We aim to measure the trade-off between the utility of data for activity recognition and privacy, e.g.~keeping gender secret. To this end, we first compare the accuracy of activity recognition and gender classification when a trained MTCNN has access to original data and to the corresponding transformed data. Then we try to measure the amount of sensitive information which is still available in the transformed data using different methods.

\begin{table}[t!]
	{
	\centering
	\begin{tabular}{|p{2.3cm}|p{1cm}|p{1.6cm}|p{0cm}}\cline{1-3}
		& \textbf{MobiAct } & \textbf{MotionSense } &\\\cline{1-3}
	 	\textbf{\#Males} & \centering{32} & \centering{14} & \\\cline{1-3}
		\textbf{\#Females} & \centering{16} & \centering{10} &\\\cline{1-3}
		\textbf{\#Features ($m$)} & \centering{9} & \centering{12} &\\ \cline{1-3}
		\textbf{Sample~Rate~(Hz)} & \centering{20} &\centering{50} & \\ \cline{1-3}
	\end{tabular}
   \caption{{
		Details of the MobiAct and MotionSense datasets.}
	}
	\label{tab:dataset}
}
\end{table}

\begin{table}[t!]
	{
		\centering
		\begin{tabular}{|p{0.9cm}|p{5.2cm}|}
			\hline
			\textbf{Model}& \textbf{Layer (Neurons $|$ Kernel $|$ Chance) }  \\
			\hline
			&
			Inp($m, d$) \\ 
			& Conv(50: $1\times5$); Conv(50: $1\times3$)    \\ 
			& Dense(50);~MP($1\times2$);~DO(0.2)    \\ 
			& Conv(40: $1\times5$)  \\
			\textbf{MTCNN} & Dense(40);~MP($1\times3$);~DO(0.2)     \\ 
			& Conv(20: $1\times3$);~DO(0.2)   \\
			& Flatten;~Dense(400);~DO(0.4)    \\ 
			& OutA = Softmax(4); OutG = Sigmoid     \\ 
			\hline
			& Inp($|x|$);~Dense($|x|/2$);~Dense($|x|/4$)\\ 
			\textbf{AE} & Dense($|x|/8$)\\ 
			& Dense($|x|/4$);~Dense($|x|/2$);~Out($|x|$)\\ \hline
			
		\end{tabular}
		\caption{{Structure of the hidden layers. The activation function for all the layers is ``ReLU''. Key -- MP: MaxPooling; DO: DropOut; $|x|=m\times d$.}}
		\label{tab:nn_arc}
	}
\end{table}

\subsection{Datasets}

We use two real-world datasets: MobiAct\footnote{publicly available at: \\ \url{http://www.bmi.teicrete.gr/index.php/research/mobiact}} and MotionSense\footnote{publicly available at: \\
\url{http://github.com/mmalekzadeh/motion-sense}}. The latter dataset is one of the contributions of this paper.

\textbf{MobiAct~\cite{vavoulas2016mobiact}} includes accelerometer, gyroscope and orientation data ($m=9$) from a smartphone collected when data subjects performed 9 activities in 16 trials. A total of 67 participants in a range of gender, age, weight, and height collected the data with a Samsung Galaxy S3 smartphone (we use a subset of 48 subjects who have no missing data). Unlike other datasets, which require the smartphone to be rigidly placed on the human body and with a specific orientation, MobiAct attempted to simulate every-day usage of mobile phones where a smartphone is located with random orientation in a loose pocket chosen by the subject (Table~\ref{tab:dataset}). 

\textbf{MotionSense} includes the accelerometer (acceleration and gravity), attitude (pitch, roll, yaw) and gyroscope data ($m=12$)  collected with an iPhone 6s kept in the participant's front pocket using SensingKit~\cite{katevas2014poster}. A total of 24 participants in a range of gender, age, weight, and height performed 6 activities in 15 trials in the same environment and conditions: downstairs, upstairs, walking, jogging, sitting, and standing. With this dataset, we aim to look for {\em personal attributes fingerprints} in time-series of sensor data, i.e.~attribute-specific patterns that can be used to infer physical and demographic attributes of the data subjects in addition to their activities. 

See \url{http:github.com/mmalekzadeh/motion-sense} for details on the methodology and the data~(Table~\ref{tab:dataset}).\\

\subsection{Experimental Setup}
For each dataset, we consider two types of setting, namely Trial and Subject. In {\em Trial}, we keep $2/3$ of trials for training and $1/3$ of them for testing. For example, if there are 3 walking trials per participant, we keep the first two trials for training and the last one for testing. In {\em Subject} we keep data of 75\% of all subjects for training and the data of remaining 25\% subjects for testing. In the Subject setting, we report the average results of four selections for test dataset.  

We train an MTCNN as the Estimator by considering two tasks: (i) activity recognition (4 classes) with \textit{categorical cross-entropy} loss function~\cite{chollet2015keras}, and (ii) gender classification (2 classes) with \textit{binary cross-entropy} loss function.~\cite{chollet2015keras}. After training MTCNN, we freeze the weights of the MTCNN layers and attach the output of a deep autoencoder~(AE) as the Guardian to the input of the MTCNN to build the GEN neural network. Finally, we compile GEN and set its loss function equals to the objective function of the Neutralizer in Equation~(\ref{eq:neutralizer}). The deep network architectures are  described in Table~\ref{tab:nn_arc}.\\

\subsection{Transformation Efficiency}
Table~\ref{tab:acc} shows that the Guardian produces time-series that keep the utility of non-sensitive inferences at a comparable level to the original ones (the average loss is three percentage points) while preventing sensitive inferences, as the gender classification accuracy decreases from more than 90\% to near the target random guess (50\%). \newline

\begin{table}[t!]
		{
	\centering
\begin{tabular}{|p{0.6cm}|p{1.4cm} |p{0.15cm}|p{0.3cm}|p{0.3cm}|}
	\hline
	Setting&Dataset&Inf.& $S_d$ &$\hat{S}_d$ \\ 
	\hline
	\multirow{4}{*}{\textbf{Trial}}&\multirow{2}{*}{\textbf{MotionSense}}& $I_{\textbf{a}}$  \cellcolor{green!10}& 95.08 \cellcolor{green!10} & 93.71  \cellcolor{green!10}  \\
	&&  $I_{\textbf{g}}$  \cellcolor{red!10} & \cellcolor{red!10} 95.15	 &\cellcolor{yellow!30} 49.32 \\
	\cline{2-3}
	&\multirow{2}{*}{\textbf{MobiAct}}&  $I_{\textbf{a}}$\cellcolor{green!20}& 94.31  \cellcolor{green!20} &90.46\cellcolor{green!20}\\
	&&   $I_{\textbf{g}}$  \cellcolor{red!10}&93.74 \cellcolor{red!10}& 49.83\cellcolor{yellow!30} \\
	\cline{1-5}
	\multirow{4}{*}{\textbf{Subject}}&\multirow{2}{*}{\textbf{MotionSense}}& $I_{\textbf{a}}$ \cellcolor{green!20} &86.33 \cellcolor{green!20} &85.19 \cellcolor{green!20}  \\
	&&  $I_{\textbf{g}}$  \cellcolor{red!10} &75.35 \cellcolor{red!10} &\cellcolor{yellow!30}52.16 \\
	\cline{2-5}
	&\multirow{2}{*}{\textbf{MobiAct}}&  $I_{\textbf{a}}\cellcolor{green!20}$ &\cellcolor{green!20}70.49&\cellcolor{green!20}65.01  \\
		&&  $I_{\textbf{g}}$ \cellcolor{red!10} &\cellcolor{red!10}66.18&45.54\cellcolor{yellow!30} \\
		\hline
	\end{tabular}
	\caption{{Activity recognition, $I_{\textbf{a}}$,  and gender classification, $I_{\textbf{g}}$, accuracy for \textit{original}, $S_d$, and \textit{transformed}, $\hat{S}_d$, data in percent (\%).}}
	\label{tab:acc}
}
\end{table}

{\bf Cross-Dataset Validation}. We also validate GEN in an ecosystem where edge users benefit from pre-trained models of a service provider. At the {\em cloud side} the Estimator (MTCNN) is trained on a public dataset, the MobiAct dataset in our case. At the \textit{edge side}, the Guardian receives the trained Estimator and uses its locally (personally) defined Neutralizer to transform the user's data, the MotionSense dataset in our case.  

The results show that the accuracy of the Estimator on raw data for $I_a$ and $I_g$  are $93.67\%$ and $92.80\%$, respectively; whereas on transformed data are $90.92\%$ and $51.93\%$, respectively. This shows an interesting property of GEN which makes it more applicable to deploy in edge devices.

The only concern here is whether users trust the pre-trained Estimator received from an untrusted service provider. User can verify the Estimator by running it on a publicly available dataset. We leave more investigation on this concern for future work.\\

\begin{figure}[t!]
	\centering
	\includegraphics[scale=0.4]{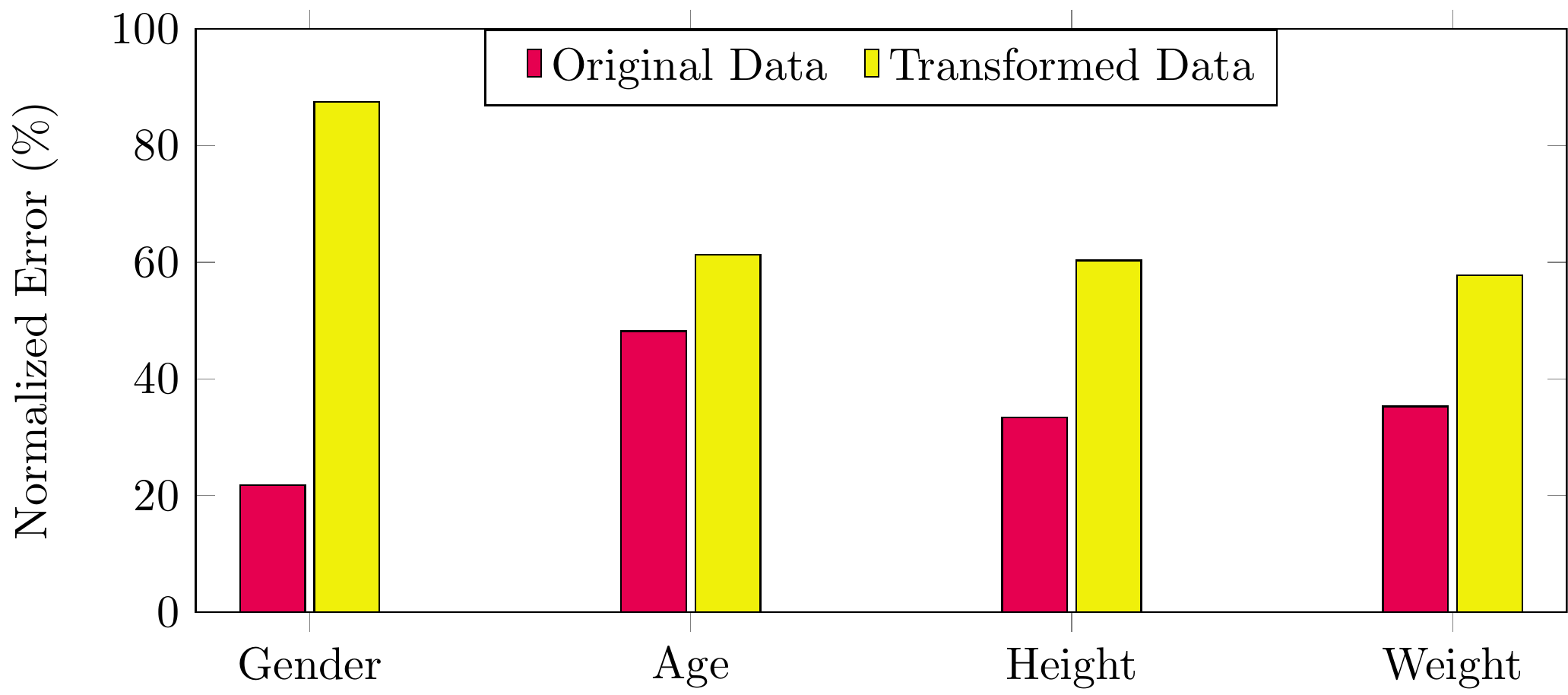}
	\caption{\label{fig:nee} {Error for gender is ``classification error'' and for the rest of attributes is ``mean absolute error''. All the values are divided by the error of a random estimator on the MotionSense dataset.}}
\end{figure}

\subsection{Measuring Information Leakage}
We aim to experimentally quantify the amount of information about user's attributes that is still available in the transformed data.\\

{\bf Using Dynamic Time Warping}. To measure the amount of residual attribute-information in sensor data, we chose\footnote{k-NN with DTW outperforms other methods in time-series classification, except when considerable computation and implementation cost is acceptable for very small improvements~\cite{bagnall2017great}.} k-Nearest Neighbors (k-NN) with Dynamic Time Warping (DTW)~\cite{salvador2007toward}.  We aim to verify whether a different algorithm will also fail to guess gender, even when adversaries get access to the entire time-series, and not just a section of it. To this end we build an $n\times n$  matrix $D_l$, where $n$ is the number of subjects in the dataset. For each activity $a_{l}\in\{downstairs, upstairs, walking, jogging\}$, let $d_l(i,j)$ be the distance between the time-series of users $u_i$ and  $u_j$  calculated by FastDTW~\cite{salvador2007toward}. Then, we calculate the final {\it distance} matrix $D$ as the element-wise average of all the matrices $D_l$; $d(i,j) =\frac{1}{4}\sum_{l} d_l(i,j)$. 

We calculate distance matrices $D$ and $\hat{D}$ for the original time-series and the transformed series (the output of the Guardian) respectively. Then we {\it compare} the ability of the estimation based on these matrices. For each user $u_{i}$; $i\in\{1,\dots,n\}$ (one out-of-sample), we {\it estimate} the value of each attribute $v_{a}(u_i)$;   $a\in\{gender, age, weight, height\}$, using distance weighted k-NN  based on matrix $D$, where the weight is:
$$
		w(i,j) = \frac{1}{d(i,j)^{2}}.
$$

Figure~\ref{fig:nee} shows that the estimation error for gender classification approaches that of a random estimator after transformation. In this Figure, the error of a random estimator for gender is $\frac{N_f}{N_f+N_m}=\frac{10}{24}$ and for the rest of attributes is considered as the half of the variation interval in dataset; e.g. $\frac{190-161}{2}=14.5$ for height.

Thus the GEN eliminates similarities between same-gender time-series and an attacker cannot confidently use distance measures to make inference about gender. Interestingly, by eliminating gender information, we also partially eliminate information on other attributes, as there are dependencies between attributes. For example, the estimation error for height and weight increases by near 25\% and 20\%, respectively. 

\begin{figure}[t!]
	\centering
	\includegraphics[scale=0.35]{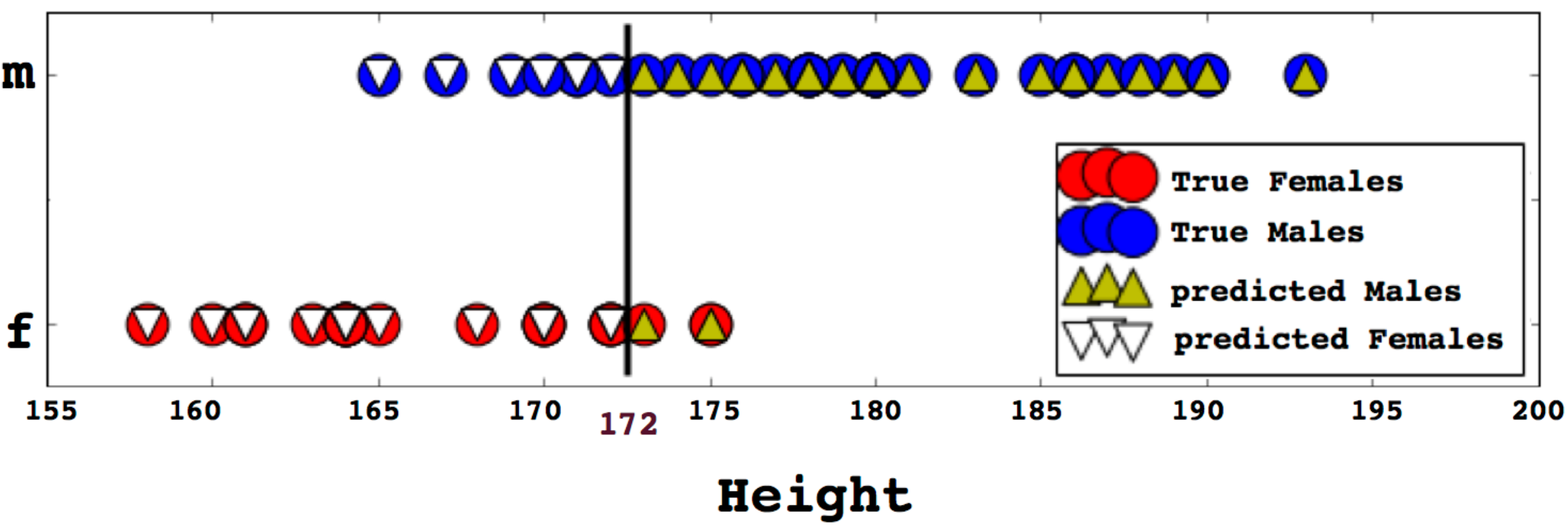}
	\caption{\label{fig:gen_hei} {Dependencies between height and gender on the MotionSense and MobiAct datasets. A classification threshold of $172cm$ predicts gender with 84\% accuracy.}}
\end{figure}

Height is indeed highly correlated with gender in both datasets (Figure~\ref{fig:gen_hei}): the prediction accuracy of gender-based on height only is 81\%.  However, gender prediction from both datasets using the MTCNN architecture is considerably better than that.\\

\begin{figure}[t!]
	\centering
	\includegraphics[scale=0.44]{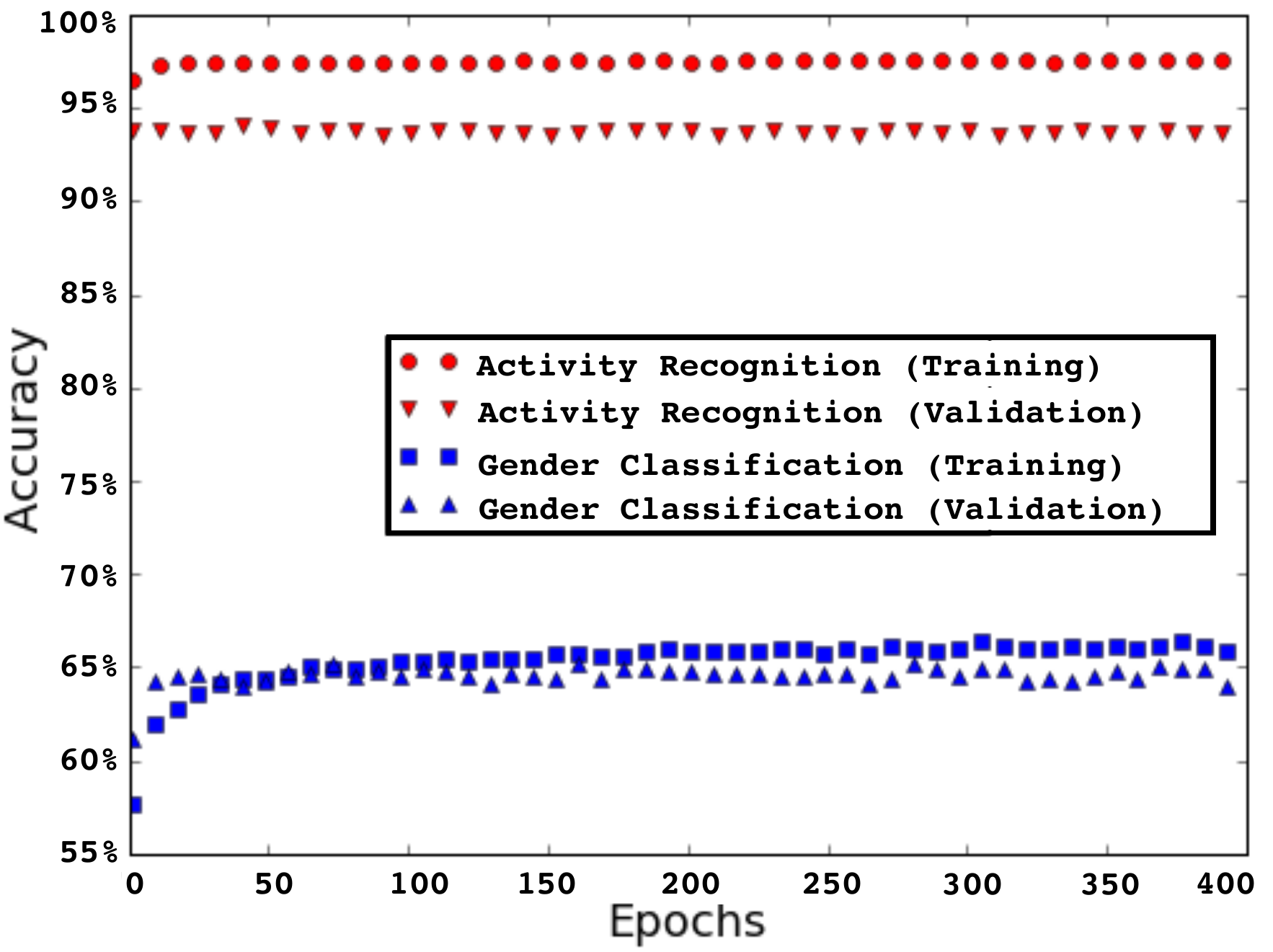}
	\caption{\label{fig:dsl} {Activity and gender classification accuracy, on the MotionSense dataset in Trial setting, when the Estimator is trained on transformed data produced by the Guardian. Although activity-features can be easily learned, there is no useful discerning information about gender.}}
\end{figure}

{\bf Using Supervised Learning}. We  explore learning gender discriminative features from {\em transformed} data. Figure~\ref{fig:dsl} shows the training and validation accuracy of activity recognition and gender classification using supervised learning on transformed data. Gender-discriminative features in the transformed data are rare, even with a large number of epochs as in this experiment. GEN eliminates gender-related features and thus makes it is difficult for a classifier to train on them even when it has access to the labels of transformed data.

Although, with experiments in this section, we have shown an acceptable efficiency in eliminating sensitive information, it is highly desired to statistically prove the efficiency of the proposed solution. Generally, high temporal granularity of time-series and strong correlation between their samples make this task very challenging. We leave exploring this area to future research.\\
\section{Related Work and Discussion}

Generative adversarial networks (GANs)~\cite{goodfellow2014generative} learn to capture the statistical distribution
of data for synthesizing new samples from the learned distribution. In the GANs a discriminator model learns to determine whether a sample is from the model distribution (i.e.~from the generator) or from the data distribution (i.e.~from a real-world source). The discriminator aims to maximize an objective function in minimax game that the generator aims to minimize. 
GANs have also been applied for enhancing privacy~\cite{huang2017context, tripathy2017privacy}. For example, to protect health records, synthetic medical datasets can be published instead of the real ones using generative models training on sensitive real-world medical datasets~\cite{choi2017generating,esteban2017real}. To provide a formal privacy guarantee, \cite{beaulieu2017privacy} trains GANs under the constraint of differential privacy~\cite{dwork2008differential} to protect against common privacy attacks. 

Although the architecture of our proposed framework looks similar to GANs, there are key structural and logical differences with other existing frameworks. First, the focus of existing works is mainly on protecting users' privacy against membership attack by releasing a synthetic dataset through differential privacy constraints. Instead, we consider a situation where a user wants to grant third parties access to sensor data that can be used to make both sensitive and non-sensitive inferences. 

Second, the generator in GANs seeks to learn the underlying distribution of the data to produce realistic simulated samples from random vectors. Instead, the Guardian in GEN seeks to partition the underlying features of the data to reconstruct privacy-preserving outputs from real-world input vectors.

Finally, the minimax game in GANs is a two-player game between generator and discriminator (i.e.~two models) that updates weights of both models in each iteration. Instead the minimax objective of GEN is a trade-off between utility and privacy that updates the weights of one only model (i.e.~the guardian) in each iteration. 

Previous works on data collected from embedded sensors of personal devices, such as~\cite{malekzadeh2017replacement, saleheen2016msieve},  consider temporal inferences on different activities over time (i.e.~some sections of time-series corresponding to non-sensitive activities and some of them to sensitive ones). In this paper, for the first time, we concurrently consider both activity and attribute inferences on the same section of time-series.

Our framework is  applicable in distributed environments: we have shown that the Estimator can be trained remotely (e.g.~on a powerful system and with a large dataset) and edge devices just need to download the resulting trained model to use it as the Estimator part of their locally implemented GEN under user's control. For example, the Guardian can be trained in user side using individuals' personal data processing platforms, like Databox~\cite{haddadi2015personal}.\\

\section{Conclusion}

We proposed the GEN framework for locally transforming sensor data on mobile edge devices to respect functions and requirements of an application as well as user privacy. We evaluated the efficiency of the trade-off between utility and privacy GEN provides on real-world datasets of motion data. 

Open questions to be explored in future work include providing theoretical bounds on the amount of sensitive information leakage after transformation and exploring dependencies between different attributes, e.g.~co-dependence of gender and height. Finally, we will measure the costs and requirements for running GEN on edge devices.\\

\begin{acks}
This work was kindly supported by the Life Sciences
Initiative at Queen Mary University London and a Microsoft
Azure for Research Award. Hamed Haddadi was partially
funded by EPSRC Databox grant (Ref: EP/N028260/1).\\
\end{acks}
	
\bibliographystyle{ACM-Reference-Format}

\begin{thebibliography}{17}
	\providecommand{\natexlab}[1]{#1}
	\providecommand{\url}[1]{\texttt{#1}}
	\expandafter\ifx\csname urlstyle\endcsname\relax
	\providecommand{\doi}[1]{doi: #1}\else
	\providecommand{\doi}{doi: \begingroup \urlstyle{rm}\Url}\fi
	
	\bibitem[Bagnall et~al.(2017)Bagnall, Lines, Bostrom, Large, and
	Keogh]{bagnall2017great}
	A.~Bagnall, J.~Lines, A.~Bostrom, J.~Large, and E.~Keogh.
	\newblock The great time series classification bake off: a review and
	experimental evaluation of recent algorithmic advances.
	\newblock \emph{Data Mining and Knowledge Discovery}, 31\penalty0 (3):\penalty0
	606--660, 2017.
	
	\bibitem[Beaulieu-Jones et~al.(2017)Beaulieu-Jones, Wu, Williams, and
	Greene]{beaulieu2017privacy}
	B.~K. Beaulieu-Jones, Z.~S. Wu, C.~Williams, and C.~S. Greene.
	\newblock Privacy-preserving generative deep neural networks support clinical
	data sharing.
	\newblock \emph{bioRxiv}, page 159756, 2017.
	
	\bibitem[Choi et~al.(2017)Choi, Biswal, Malin, Duke, Stewart, and
	Sun]{choi2017generating}
	E.~Choi, S.~Biswal, B.~Malin, J.~Duke, W.~F. Stewart, and J.~Sun.
	\newblock Generating multi-label discrete electronic health records using
	generative adversarial networks.
	\newblock \emph{arXiv preprint arXiv:1703.06490}, 2017.
	
	\bibitem[Chollet et~al.(2015)]{chollet2015keras}
	F.~Chollet et~al.
	\newblock Keras.
	\newblock \url{https://github.com/fchollet/keras}, 2015.
	
	\bibitem[Dwork(2008)]{dwork2008differential}
	C.~Dwork.
	\newblock Differential privacy: A survey of results.
	\newblock In \emph{International Conference on Theory and Applications of
		Models of Computation}, pages 1--19. Springer, 2008.
	
	\bibitem[Esteban et~al.(2017)Esteban, Hyland, and R{\"a}tsch]{esteban2017real}
	C.~Esteban, S.~L. Hyland, and G.~R{\"a}tsch.
	\newblock Real-valued (medical) time series generation with recurrent
	conditional gans.
	\newblock \emph{arXiv preprint arXiv:1706.02633}, 2017.
	
	\bibitem[Goodfellow et~al.(2014)Goodfellow, Pouget-Abadie, Mirza, Xu,
	Warde-Farley, Ozair, Courville, and Bengio]{goodfellow2014generative}
	I.~Goodfellow, J.~Pouget-Abadie, M.~Mirza, B.~Xu, D.~Warde-Farley, S.~Ozair,
	A.~Courville, and Y.~Bengio.
	\newblock Generative adversarial nets.
	\newblock In \emph{Advances in neural information processing systems}, pages
	2672--2680, 2014.
	
	\bibitem[Haddadi et~al.(2015)Haddadi, Howard, Chaudhry, Crowcroft,
	Madhavapeddy, McAuley, and Mortier]{haddadi2015personal}
	H.~Haddadi, H.~Howard, A.~Chaudhry, J.~Crowcroft, A.~Madhavapeddy, D.~McAuley,
	and R.~Mortier.
	\newblock Personal data: thinking inside the box.
	\newblock In \emph{Proceedings of The Fifth Decennial Aarhus Conference on
		Critical Alternatives}, pages 29--32. Aarhus University Press, 2015.
	
	\bibitem[Huang et~al.(2017)Huang, Kairouz, Chen, Sankar, and
	Rajagopal]{huang2017context}
	C.~Huang, P.~Kairouz, X.~Chen, L.~Sankar, and R.~Rajagopal.
	\newblock Context-aware generative adversarial privacy.
	\newblock \emph{Entropy}, 19\penalty0 (12):\penalty0 656, 2017.
	
	\bibitem[Katevas et~al.(2014)Katevas, Haddadi, and
	Tokarchuk]{katevas2014poster}
	K.~Katevas, H.~Haddadi, and L.~Tokarchuk.
	\newblock Poster: Sensingkit: A multi-platform mobile sensing framework for
	large-scale experiments.
	\newblock In \emph{Proceedings of the 20th Annual International Conference on
		Mobile Computing and Networking}, pages 375--378. ACM, 2014.
	
	\bibitem[Malekzadeh et~al.()Malekzadeh, Clegg, and
	Haddadi]{malekzadeh2017replacement}
	M.~Malekzadeh, R.~G. Clegg, and H.~Haddadi.
	\newblock Replacement autoencoder: A privacy-preserving algorithm for sensory
	data analysis.
	\newblock \emph{The 3rd ACM/IEEE International Conference on Internet-of-Things
		Design and Implementation, 2018}.
	
	\bibitem[Saleheen et~al.(2016)Saleheen, Chakraborty, Ali, Rahman, Hossain,
	Bari, Buder, Srivastava, and Kumar]{saleheen2016msieve}
	N.~Saleheen, S.~Chakraborty, N.~Ali, M.~M. Rahman, S.~M. Hossain, R.~Bari,
	E.~Buder, M.~Srivastava, and S.~Kumar.
	\newblock msieve: differential behavioral privacy in time series of mobile
	sensor data.
	\newblock In \emph{Proceedings of the 2016 ACM International Joint Conference
		on Pervasive and Ubiquitous Computing}, pages 706--717, 2016.
	
	\bibitem[Salvador and Chan(2007)]{salvador2007toward}
	S.~Salvador and P.~Chan.
	\newblock Toward accurate dynamic time warping in linear time and space.
	\newblock \emph{Intelligent Data Analysis}, 11\penalty0 (5):\penalty0 561--580,
	2007.
	
	\bibitem[Tripathy et~al.(2017)Tripathy, Wang, and Ishwar]{tripathy2017privacy}
	A.~Tripathy, Y.~Wang, and P.~Ishwar.
	\newblock Privacy-preserving adversarial networks.
	\newblock \emph{arXiv preprint arXiv:1712.07008}, 2017.
	
	\bibitem[Vavoulas et~al.(2016)Vavoulas, Chatzaki, Malliotakis, Pediaditis, and
	Tsiknakis]{vavoulas2016mobiact}
	G.~Vavoulas, C.~Chatzaki, T.~Malliotakis, M.~Pediaditis, and M.~Tsiknakis.
	\newblock The mobiact dataset: Recognition of activities of daily living using
	smartphones.
	\newblock In \emph{ICT4AgeingWell}, pages 143--151, 2016.
	
	\bibitem[Vincent et~al.(2008)Vincent, Larochelle, Bengio, and
	Manzagol]{vincent2008extracting}
	P.~Vincent, H.~Larochelle, Y.~Bengio, and P.-A. Manzagol.
	\newblock Extracting and composing robust features with denoising autoencoders.
	\newblock In \emph{Proceedings of the 25th International Conference on Machine
		learning}, pages 1096--1103, 2008.
	
	\bibitem[Yang et~al.(2015)Yang, Nguyen, San, Li, and
	Krishnaswamy]{yang2015deep}
	J.~Yang, M.~N. Nguyen, P.~P. San, X.~Li, and S.~Krishnaswamy.
	\newblock Deep convolutional neural networks on multichannel time series for
	human activity recognition.
	\newblock In \emph{Proceedings of the 24th International Conference on
		Artificial Intelligence}, pages 3995--4001, 2015.
	
\end{thebibliography}

\end{document}